\documentclass[10pt,conference]{IEEEtran}
\usepackage{cite}
\usepackage{amsmath,amssymb,amsfonts}
\usepackage{algorithm}
\usepackage{algpseudocode}
\usepackage{graphicx}
\usepackage{textcomp}
\usepackage{xcolor}
\usepackage[hyphens]{url}
\usepackage{fancyhdr}
\usepackage{hyperref}
\usepackage{multirow}
\usepackage{float}
\usepackage{listings}
\usepackage{enumitem}
\usepackage{balance}

\newcommand{\hpcayear}{2026}
\newcommand{\vectorlite}[0]{\textsc{VectorLiteRag}}
\newcommand{\worktitle}{\textsc{VectorLiteRAG}}
\renewcommand{\appendixname}{Artifact Appendix}

\newcommand{\hpcasubmissionnumber}{876}
\title{VectorLiteRAG: Latency-Aware and Fine-Grained \\ Resource Partitioning for Efficient RAG}

\def\hpcacameraready{} 

\newcommand\hpcaauthors{Junkyum Kim and Divya Mahajan}
\newcommand\hpcaaffiliation{Georgia Institute of Technology}
\newcommand\hpcaemail{\{jun-kyum.kim, divya.mahajan\}@gatech.edu}

\author{
  \ifdefined\hpcacameraready
    \IEEEauthorblockN{\hpcaauthors{}}
      \IEEEauthorblockA{
        \hpcaaffiliation{} \\
        \hpcaemail{}
      }
  \else
    \IEEEauthorblockN{\normalsize{HPCA \hpcayear{} Submission
      \textbf{\#\hpcasubmissionnumber{}}} \\
      \IEEEauthorblockA{
        Confidential Draft \\
        Do NOT Distribute!!
      }
    }
  \fi 
}

\fancypagestyle{camerareadyfirstpage}{%
  \fancyhead{}
  
  \fancyhead[C]{
    \ifdefined\aeopen
    \parbox[][12mm][t]{13.5cm}{\hpcayear{} IEEE International Symposium on High-Performance Computer Architecture (HPCA)}    
    \else
      \ifdefined\aereviewed
      \parbox[][12mm][t]{13.5cm}{\hpcayear{} IEEE International Symposium on High-Performance Computer Architecture (HPCA)}
      \else
      \ifdefined\aereproduced
      \parbox[][12mm][t]{13.5cm}{\hpcayear{} IEEE International Symposium on High-Performance Computer Architecture (HPCA)}
      \else
      \parbox[][0mm][t]{13.5cm}{\hpcayear{} IEEE International Symposium on High-Performance Computer Architecture (HPCA)}
    \fi 
    \fi 
    \fi 
    \ifdefined\aeopen 
      \includegraphics[width=12mm,height=12mm]{ae-badges/open-research-objects.pdf}
    \fi 
    \ifdefined\aereviewed
      \includegraphics[width=12mm,height=12mm]{ae-badges/research-objects-reviewed.pdf}
    \fi 
    \ifdefined\aereproduced
      \includegraphics[width=12mm,height=12mm]{ae-badges/results-reproduced.pdf}
    \fi
  }
  \fancyfoot[C]{}
}
\begin{document}
\maketitle

\ifdefined\hpcacameraready 
  \thispagestyle{camerareadyfirstpage}
  \pagestyle{empty}
\else
  \thispagestyle{plain}
  \pagestyle{plain}
\fi

\newcommand{\hpcaheight}{0mm}
\ifdefined\eaopen
\renewcommand{\hpcaheight}{12mm}
\fi


\begin{abstract}

Retrieval-Augmented Generation leverages vector similarity search to enhance large language models with up-to-date, external knowledge, enabling accurate and reliable responses. While CPU-only vector search incurs high latency on large, high-dimensional indices, co-locating the retriever and the LLM on the GPU leads to resource sharing that can create resource contention. Specifically, vector search is memory and I/O intensive, placing it in direct conflict with LLM inference, which demands memory for KV cache and compute for higher throughput. 
We present \worktitle, a latency-aware RAG serving system that explicitly orchestrates data placement and execution across retrieval and inference to meet strict end-to-end SLOs. \worktitle~is driven by access-pattern analysis and performance estimation to regulate how retrieval variability can be mitigated and managed in the system with LLM inference, enabling SLO-compliant execution under skewed and dynamic workloads.
By jointly modeling search latency and query hit-rate distributions, \worktitle~identifies an optimal index partitioning point across CPU and GPU that minimizes contention and stabilizes batching behavior, thereby maximizing sustained throughput under skewed access patterns. 
A low-overhead online index update mechanism allows \worktitle~to continuously adapt to evolving request distributions, preserving batching efficiency and throughput as access patterns evolve.
Our evaluations demonstrate that \worktitle~consistently expands the range of SLO-compliant request rate across all tested configurations. Without increasing the generation latency or requiring additional hardware, \worktitle~outperforms both naive and existing alternative frameworks, improving attainable SLO-bound throughput by up to 1.5$\times$.

\end{abstract}

\section{Introduction} \label{sec:introduction}

\par Retrieval-Augmented Generation (RAG) is a powerful system in natural language processing, particularly for domain-specific question answering and information retrieval tasks~\cite{bib:rag, bib:realm, bib:ralm, bib:ragsurvey1, bib:ragsurvey2}. Its key strength lies in combining parametric memory, encoded in the weights of a large language model, with non-parametric memory retrieved from an external knowledge corpus. Although parametric memory provides strong generalization, it is expensive to train and difficult to update. To mitigate this, RAG pipelines first perform similarity search using approximate nearest neighbor search (ANNS) algorithms to retrieve relevant documents from a large database. The retrieved documents are then fed into the LLM's context to generate up-to-date and reliable responses.

\begin{figure} 
    \centering
    \includegraphics[width=\linewidth]{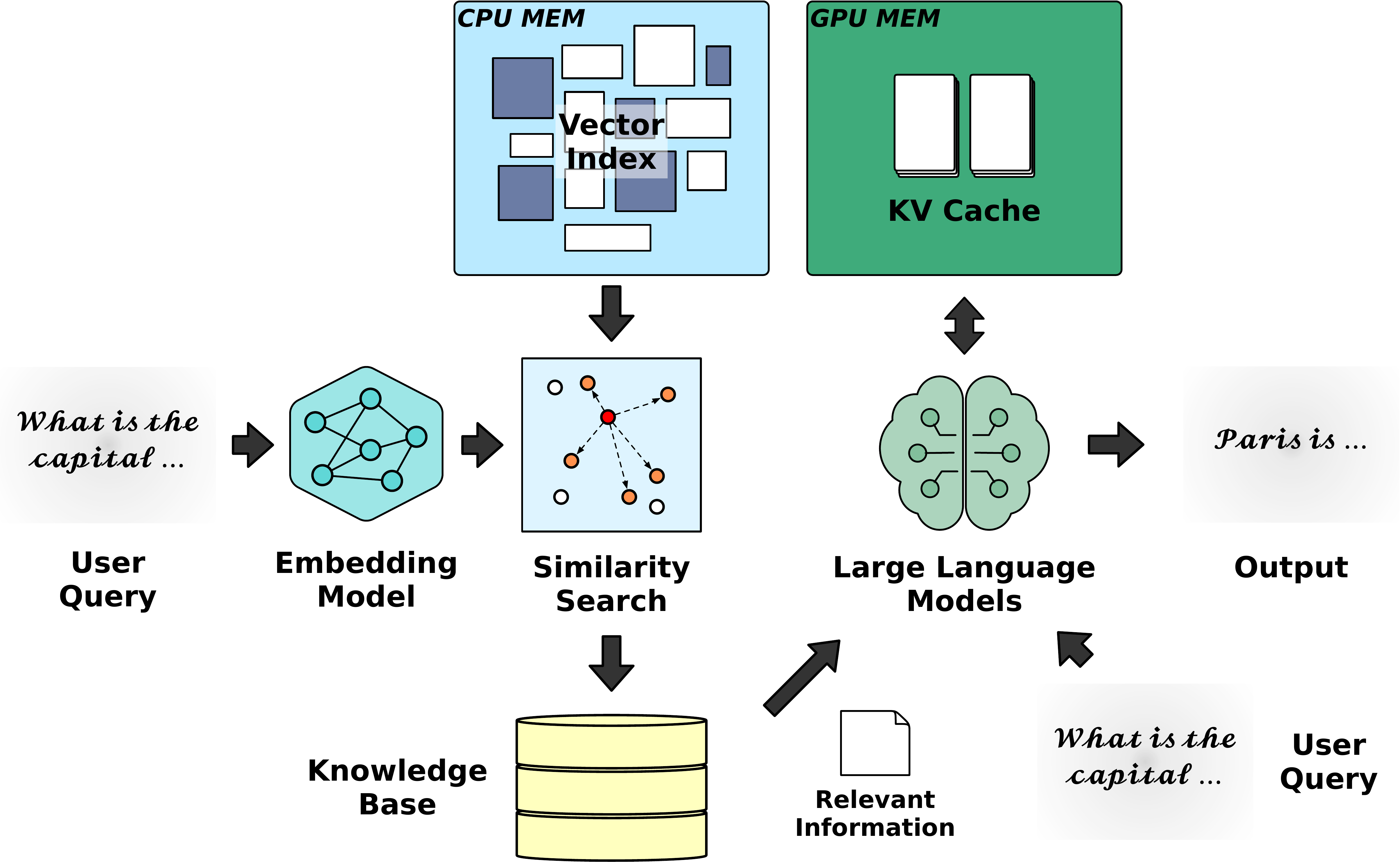} 
    \caption{End-to-end pipeline of a RAG system, where the input query is indexed into the vector database stored in memory, while the knowledge corpus resides in storage. The LLM prefill and decode execute on the GPU.}
    \label{fig:rag-pipeline}
    \vspace{-1 em}
\end{figure}

RAG frameworks~\cite{bib:langchain, bib:llamaindex, bib:milvus} typically adopt heterogeneous hardware configurations, where vector retrieval is executed on CPUs and LLM generation is served by GPUs. This is driven by the system characteristics: LLM inference requires massive matrix multiplications and benefits significantly from GPU acceleration, whereas retrieval has traditionally been seen as a lighter task suited for CPUs. Offloading retrieval to CPUs allows GPUs to be dedicated to the more compute-intensive generation phase. CPU-based vector search may be sufficient for small vector databases, however, as the dimensionality of the embeddings and the size of the dataset grow, retrieval becomes increasingly compute- and memory- bound. CPUs, with limited parallelism, narrower vector units, and lower memory bandwidth, struggle to handle high-throughput similarity search at scale. 

This latency imbalance creates a bottlenecked pipeline where the relatively slow CPU-based retrieval delays the GPU-accelerated generation phase, reducing the benefits of fast LLM inference and degrading overall system responsiveness.
In our observations, CPU-based retrieval can take up to twice as long as the LLM prefill phase, increasing the total Time-to-First-Token (TTFT) from 197ms to 606ms when using a large database with 128M vectors, compared to a language model (Llama3-8B) operating without retrieval.

Although the retrieval operation is computationally lighter than the generation phase, it can still benefit significantly from GPU acceleration for two reasons:  
(1) GPUs feature wide and powerful vector units that enable highly parallelized distance computations, offering superior performance for similarity calculations on long embedding vectors.  
(2) The retrieval process involves scanning intermediate distance tensors to identify the closest data points in the vector space. These operations are typically implemented as memory lookups a task where GPUs outperform CPUs due to their vectorized memory access and higher I/O.

\par In addition to compute and bandwidth demands, vector retrieval introduces significant memory pressure. To reduce memory footprint and speed up the search process, vector databases are commonly compressed into vector indexes using quantization techniques such as product quantization (PQ)~\cite{bib:pq}. Nevertheless, even after compression, vector indexes still occupy significant memory space, often exceeding the memory capacity of a GPU. Furthermore, intermediate data structures such as distances between cluster centroids and queries consume additional memory. 

\par These compute and memory pressures create a resource tension between the retrieval and generation stages, especially as the vector database grows and CPU-based search fails to meet strict latency requirements. GPU memory is already constrained, with most of it reserved for model weights and KV cache for the LLM. Naively sharding the vector index across all GPUs can lead to memory contention and reduced overall throughput. Alternatively, assigning a disaggregated GPU for retrieval can prevent direct interference between stages, but degrades overall system throughput by reducing the number of available LLM instances, in particular when models require multiple GPUs, enforcing rigid allocation schemes.

Motivated by these challenges, this work explores a holistic approach to optimizing distributed RAG pipelines through joint resource allocation between vector search and LLM generation. We present \worktitle, a system that partitions the vector index between GPU and CPU-based on query access patterns and LLM deployment configurations, aiming to maximize throughput while meeting latency targets by exploiting the compute power of GPUs across both stages of the RAG pipeline.
By analytically modeling similarity search latency, we determine the smallest index portion that needs to be placed on the GPU to satisfy the latency requirement under a given system configuration. Accordingly, \worktitle~offers a latency-aware, throughput-optimized solution that requires no additional hardware resources. This approach is grounded in two key insights:

\noindent \textbf{Access-Skew-Aware Data Layout.}
\worktitle~leverages a key characteristic of Inverted File (IVF) based retrieval systems~\cite{bib:ivf}, that query accesses exhibit skew across clusters. To take advantage of this, \worktitle~incorporates an analytical model that determines the optimal partitioning point and corresponding layout for a multi-GPU system. While the coarse quantizer and cold clusters remain on the CPU, a small subset of hot clusters are cached and distributed across GPUs. The system allocates just enough hot clusters to the GPUs, avoiding both oversubscription of GPU resources during retrieval.

\noindent \textbf{Inter/Intra-Query Variance-Aware Routing.}
When hot clusters are distributed across GPUs, hit rates vary both across queries (inter-query variance) and across device shards within a query (intra-query variance). Existing systems that enforce fixed retrieval configurations across devices fail to account for this variability and often over-allocate GPU threads. \worktitle~introduces query- and shard- aware routing to avoid such inefficiencies. After determining the most relevant clusters, it dispatches work to CPU or GPU based on their actual expected contribution. It also monitors per-query progress, forwarding early-finishing queries to reduce straggler-induced delays and improve batching efficiency.

\noindent Our contributions are summarized as follows:

\begin{itemize}[leftmargin=*, itemsep=1pt, parsep=0pt, topsep=1pt, partopsep=1pt]
\item \textbf{Access-skew modeling and hit-rate estimation.} We characterize access skew in IVF-based retrieval systems and develop a hit-rate estimation method based on observed cluster access patterns.

\item \textbf{Analytical latency model and SLO-aware partitioning.} We construct a latency model that accounts for inter-query variance and use it to determine the optimal CPU-GPU index partitioning point that meets latency targets.

\item \textbf{Distributed retrieval pipeline.} We design a distributed retrieval pipeline that adaptively allocates search tasks across CPUs and GPUs by exploiting inter-device hit rate variance, improving efficiency and avoiding unnecessary GPU resource usage.
\end{itemize}

\section{Retrieval Augmented Generation} \label{sec:RAG}

\begin{figure}
    \centering
    \includegraphics[width=0.9\linewidth]{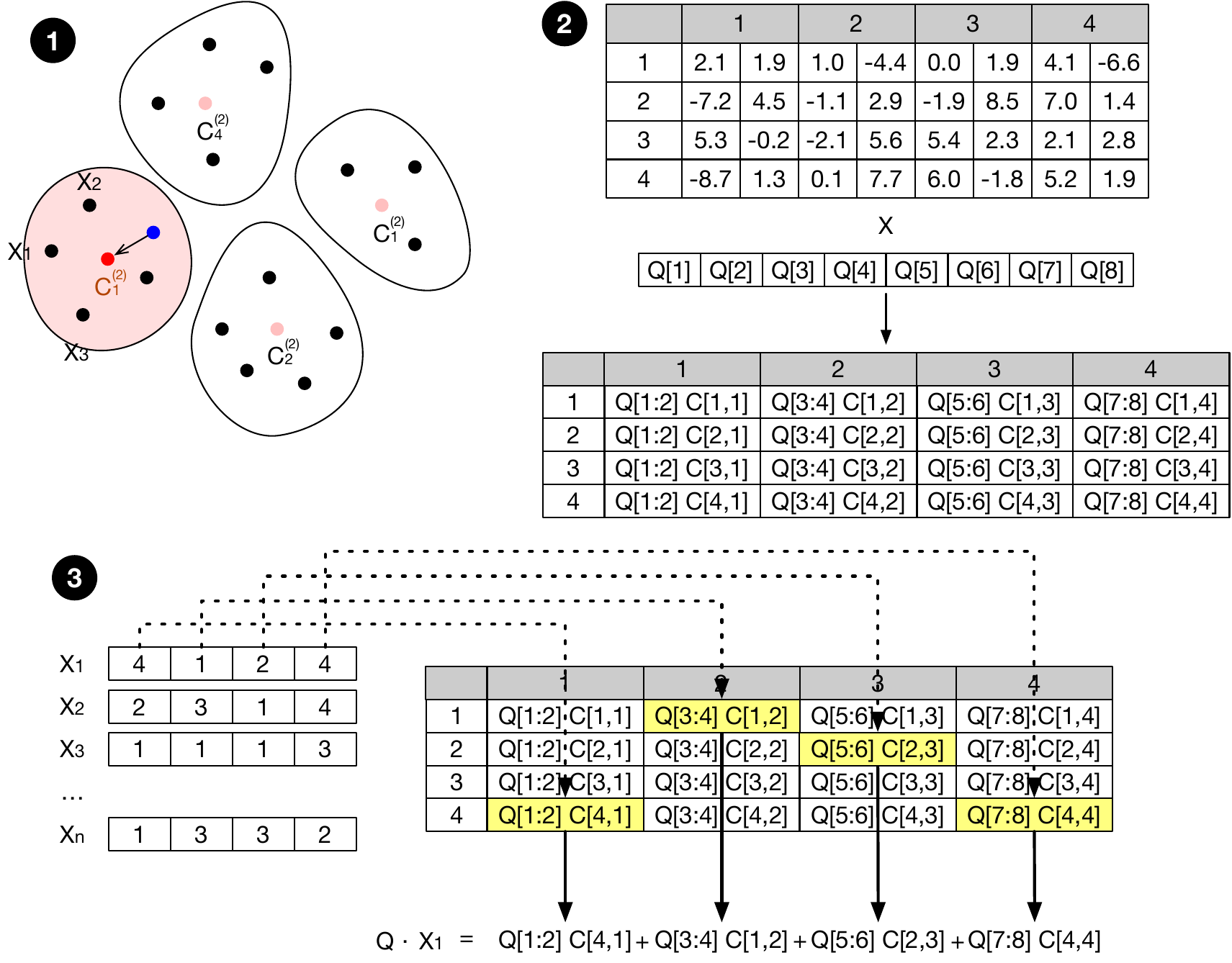}
    \caption{Three stages of vector search in IVF–based index: (1) coarse quantization to identify clusters most semantically similar to the query, (2) construction of a LUT containing partial distances between the query and codewords, and (3) scanning the LUT and re-ranking candidates from the selected clusters based on aggregated distances.}
    \label{fig:ivfpq}
    \vspace{-1 em}
\end{figure}

In a RAG system, user queries are first transformed into vector embeddings using embedding models~\cite{bib:sentence-transformer, bib:openai_embed,bib:stella,bib:mpnet}. These embeddings capture the semantics of the input and enable similarity search by comparing query vectors to a vector database constructed from the knowledge corpus, typically encoded using the same embedding model. State-of-the-art embedding models produce vectors of several thousand dimensions for higher quality, but this increased dimensionality raises the cost of distance computations.

Since exhaustive pairwise search is computationally infeasible at scale, large vector retrieval relies on approximate nearest neighbor search to efficiently identify relevant documents. The retrieved vectors are mapped back to their corresponding documents, which are provided as additional context to the LLM alongside the original query.

\begin{figure} 
    \centering
    \includegraphics[width=\linewidth]{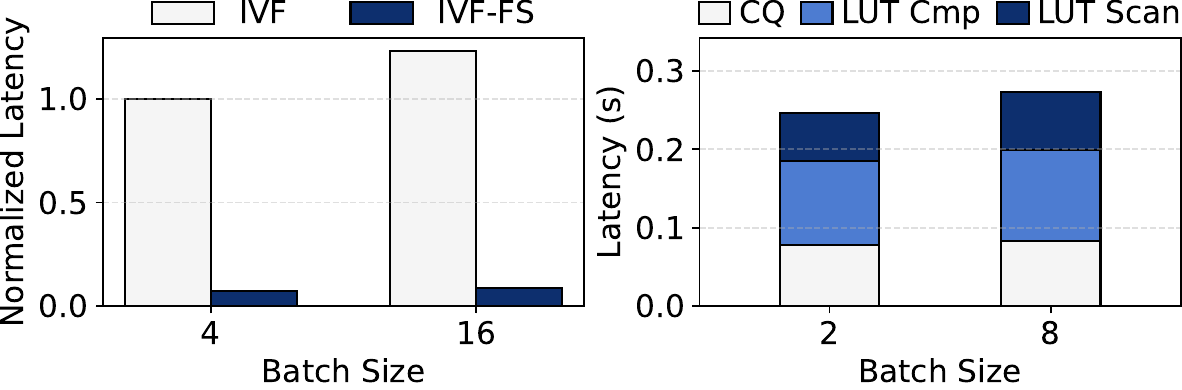} 
    \vspace{-2 em}
    \caption{\textbf{Left}: Search latency comparison between standard IVF and IVF with fast scan (IVF-FS). Except for the fast scan optimization, both indexes share identical configurations. IVF-FS achieves significantly faster search speed. \textbf{Right}: Latency breakdown of IVF-FS on a 128M vector index. Lookup table operations dominate the overall search time.}
    \label{fig:intro_breakdown}
\vspace{-1 em}
\end{figure}

\subsection{Inverted List Index IVF} \label{subsec:IVF}

There are several approaches for structuring a vector database into a searchable index. Among them, HNSW and IVF are the most widely used.

HNSW~\cite{bib:hnsw} (Hierarchical Navigable Small World) is a graph-based structure where each vector forms a node connected to its nearest neighbors. It enables rapid search via hierarchical traversal and offers fast index construction. However, the additional edge information significantly increases memory usage as the dataset grows.

In contrast, the Inverted File (IVF) index~\cite{bib:ivf} organizes the index as a hierarchical clustering structure. A subset of vectors is first clustered via K-Means to obtain centroids. Then, each database vector is assigned to the closest centroid, forming an inverted list. This structure narrows the search space using only centroid metadata, resulting in low memory overhead and high scalability. As such, IVF is widely adopted and studied in retrieval systems for large knowledge corpora~\cite{bib:hedrarag, bib:hermes, bib:scann, bib:faiss, bib:anna, bib:chameleon}.
To further reduce memory usage, quantization techniques are applied on top of IVF. Scalar quantization (SQ) reduces each vector element to a smaller numerical type (e.g., \texttt{float32} to \texttt{int8}), offering simplicity but limited compression. For higher compression ratios, product quantization (PQ)~\cite{bib:pq} is commonly used.

\subsection{Search Operation in IVF Index}
\label{subsec:IVFsearch}

Figure~\ref{fig:ivfpq} illustrates the search process in an IVF-PQ index, where an inverted list structure is combined with product quantization. When a query is received, the retriever first identifies the closest clusters, narrowing the search space. The number of clusters searched is controlled by the parameter \texttt{nprobe}, which trades off speed and accuracy.

Next, a distance lookup table is constructed. Since each vector is quantized into discrete sub-vector codes, each code maps to a representative value, trained and stored in the code-book. By pre-computing distances between the query vector and these representative values, the system avoids computing full distances to every vector. During the scan stage, these LUTs are used to accumulate approximate distances and retrieve the top-k nearest vectors. 

A deeper analysis of IVF search, shown in Figure~\ref{fig:intro_breakdown}, reveals that the large portion of the search time is spent on constructing and scanning the distance lookup table. This highlights the LUT stage as a key bottleneck in retrieval latency. To mitigate this overhead, fast scanning techniques~\cite{bib:fastscan} have been proposed and implemented in libraries such as Faiss~\cite{bib:faiss} and ScaNN~\cite{bib:scann}. These methods leverage SIMD instructions and CPU vector registers to accelerate distance lookup operations. By carefully organizing lookup tables and quantization codes into memory-aligned layouts, they significantly outperform conventional IVF scan routines, particularly in CPU-based environments.

Motivated by their superior latency-performance trade-off, we adopt fast scanning in our system to enable efficient and low-latency vector retrieval. However, despite the SIMD capabilities of modern CPUs, CPU-based search can still become a bottleneck, ultimately degrading the responsiveness of the end-to-end RAG system.

\section{Challenges and Opportunities in RAG serving} \label{sec:motivation}

\begin{figure} 
    \centering
    \includegraphics[width=\linewidth]{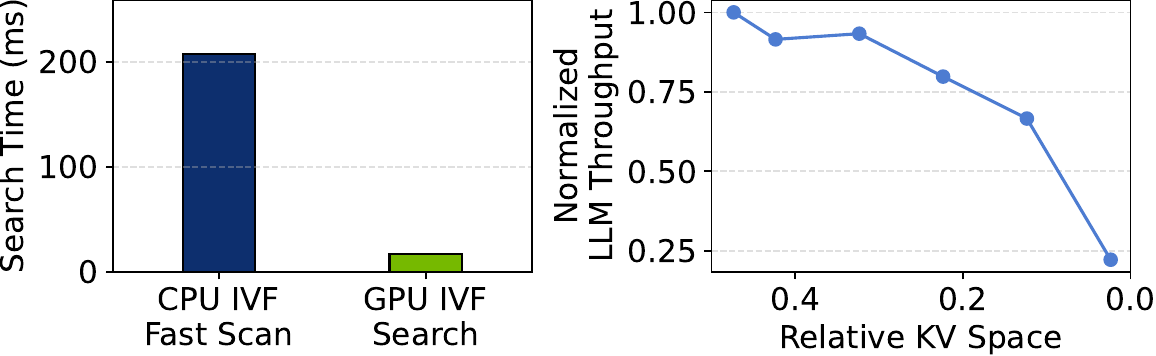} 
    \vspace{-2em}
    \caption{\textbf{Left}: While fast scanning accelerates IVF-based vector search on CPU(64 core Xeon 8462Y+), GPU(H100)-based IVF search offers superior performance. \textbf{Right}: Relationship between KV cache size and LLM throughput for the Qwen3-30B model on two H100 GPUs. Reducing KV cache space leads to a significant drop in throughput.}
    \label{fig:motiv_gpu}
\vspace{-10 pt}
\end{figure}

\subsection{GPU search vs. CPU search} \label{subsec:gpuvscpu}

While fast scan indexes significantly improve the latency of vector similarity search on CPUs, GPU-based retrieval can offer even greater speedups, due to their wider vector processing units and higher memory bandwidth. As shown in Figure~\ref{fig:motiv_gpu} (left), GPU-accelerated IVF search can outperform fast scan methods by nearly an order of magnitude.

Thus, offloading retrieval to the GPU can offer higher speedups for large-scale vector databases where CPU-based search remains a bottleneck. However, this comes with a fundamental trade-off: GPU memory is already heavily utilized by LLMs, particularly for storing KV cache and model weights. Allocating additional memory for the vector index can reduce available cache space, ultimately degrading LLM throughput, as illustrated in Figure~\ref{fig:motiv_gpu} (right).

Beyond memory capacity, GPU retrieval additionally incurs scheduling overheads due to increased contention for compute resources. Shared memory is used to stage partial distance lookup tables, and each query–cluster pair typically maps to a thread block. As the number of probed clusters increases, so does the occupancy and scheduling pressure on the GPU, further impacting performance.

\noindent \textbf{Takeaway 1. \textit{GPU-based retrieval can substantially outperform even the fastest CPU-based methods, but due to contention with LLM inference workloads, careful memory and compute allocation is essential.}}

\subsection{Opportunity of Tiered Search Structure} \label{subsec:skewinivf}
The distribution of query access patterns in IVF indexes reveals the presence of hot clusters, a small subset that dominates retrieval traffic.

\begin{figure} 
    \centering
    \includegraphics[width=\linewidth]{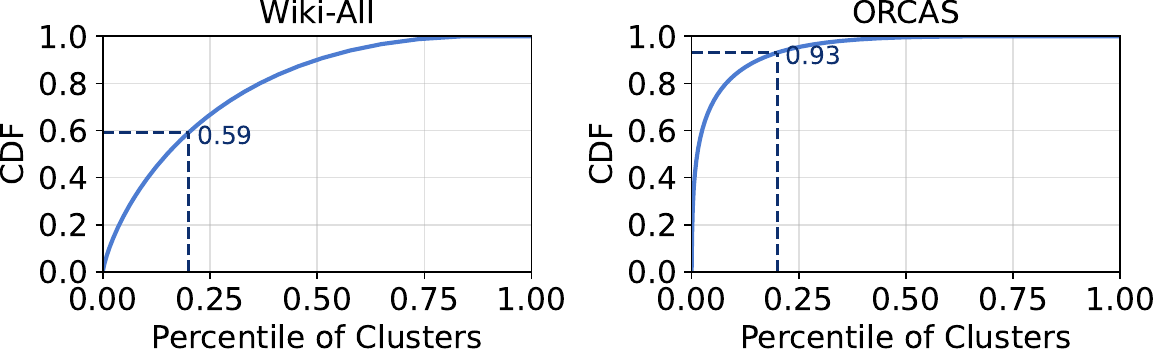} 
    \vspace{-2 em}
    \caption{CDF of cluster access frequency for queries from the Wiki-All~\cite{bib:wikiall} and ORCAS~\cite{bib:orcas} datasets. While the two distributions exhibit different levels of skewness, in both cases, the top 20\% of clusters account for over 50\% of the total distance computations.}
    \label{fig:hotclusters}
\vspace{-1.5 em}
\end{figure}

As shown on the left of Figure~\ref{fig:hotclusters}, the cumulative distribution of coarse quantization results exhibits a strong skew: the top 20\% of clusters account for nearly 60\% of accesses in Wiki-All~\cite{bib:wikiall} and over 93\% in ORCAS~\cite{bib:orcas}. This skew is especially pronounced in ORCAS, which reflects real-world query behavior through unfiltered click-through logs, capturing both popularity bias and the imbalance introduced by k-means quantization.

This imbalance results in inefficient memory usage, as significant resources are allocated to rarely accessed clusters with limited contribution to retrieval quality.

\noindent \textbf{Takeaway 2. \textit{IVF index access patterns are highly skewed: a small number of clusters account for the vast majority of retrievals. This motivates a tiered index design, where frequently accessed clusters are prioritized for acceleration (e.g., GPU caching), and cold clusters are offloaded to lower-tier compute and storage.}}

Embedding access patterns in recommendation systems are also known to exhibit significant skew, where a small subset of items or users dominates embedding lookup frequency. This observation has motivated several tiered architecture designs that prioritize popular embeddings for faster access~\cite{bib:fae, bib:hotline, bib:rhurecsys, bib:recsys2_hotcold, bib:slipstream, bib:adrec, bib:slipstream}.
Inspired by this insight, our work offers tiered acceleration to vector similarity search. However, a key distinction lies in the granularity of memory accesses. In recommendation systems, embedding look-ups are performed via embedding IDs. In contrast, vector similarity search systems conduct fully content-based retrieval, where relevant vectors must be located by computing distances to hundreds or thousands of candidates per query. To identify the nearest vector, the search must access not only the target vector but also neighboring vectors within the cluster.

Moreover, even if each embedding is uniformly accessed, clusters can contain varying numbers of vectors, exacerbating the access skew. This imbalance causes certain clusters to dominate query traffic, creating hot regions in memory access. As a result, skew in our setting emerges more prominently at the cluster level rather than the vector level.

Consequently, although both domains benefit from tiered designs, the unit of optimization and the manifestation of skew differ substantially. Our approach explicitly targets cluster-level skew in large-scale retrieval workloads, enabling effective tiered placement and latency-aware resource allocation that are not directly addressed by prior embedding-centric designs.

\begin{figure}
    \centering
    \includegraphics[width=\linewidth]{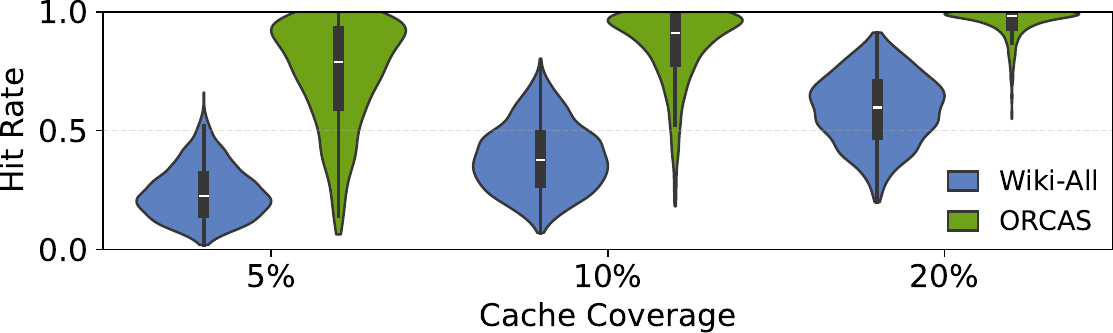} 
    \vspace{-2 em}
    \caption{Violin plot of hit rate distribution at different cache-coverages. The width of the violin indicates the density of queries with similar hit rates, while the white dot and black bar denote the median and inter-quartile range, respectively. This highlights that increasing cache coverage improves overall hit rates but does not eliminate tail queries with poor hit rates.}
    \label{fig:hitrate_variance}
\vspace{-1.5 em}
\end{figure}

\subsection{Variance of Hit Rate across Queries} \label{sec:queryvar}
While tiered resource allocation strategies can accelerate vector search by caching frequently accessed clusters, their effectiveness in deployment is often hindered by query-level variance in hit rates. Long-tail queries with less cache hits can significantly limit the overall performance gains.

Figure~\ref{fig:hitrate_variance} presents a violin plot of hit rate distributions across queries, measured by counting the number of clusters (among the total \texttt{nprobe}) that fall within the cached hot cluster set. As cache coverage increases from 5\% to 20\% of total clusters, the average hit rate improves accordingly. However, the variance remains substantial, especially in highly skewed datasets such as ORCAS, where a long tail of queries exhibits minimal cache benefit.

This variance introduces a deployment challenge. Since vector search throughput scales with batch size, retrievers are typically deployed with batching enabled. However, in the presence of low-hit queries within a batch, the entire batch’s processing time is effectively bounded by the slowest query. As a result, even if the average per-query latency is reduced by GPU acceleration, end-to-end latency improvements are constrained. Therefore, to fully realize the benefits of tiered or cached retrieval in real-world deployments, it is essential to account for such hit rate variance and long-tail behavior during system design.

\noindent \textbf{Takeaway 3. \textit{Variance in hit rate across queries poses a challenge in latency-critical deployments, due to long-tail queries as batching amplifies the impact of long-tail queries, limiting the effectiveness of caching.}}

\begin{figure*}
    \centering
    \includegraphics[width=1.0\linewidth]{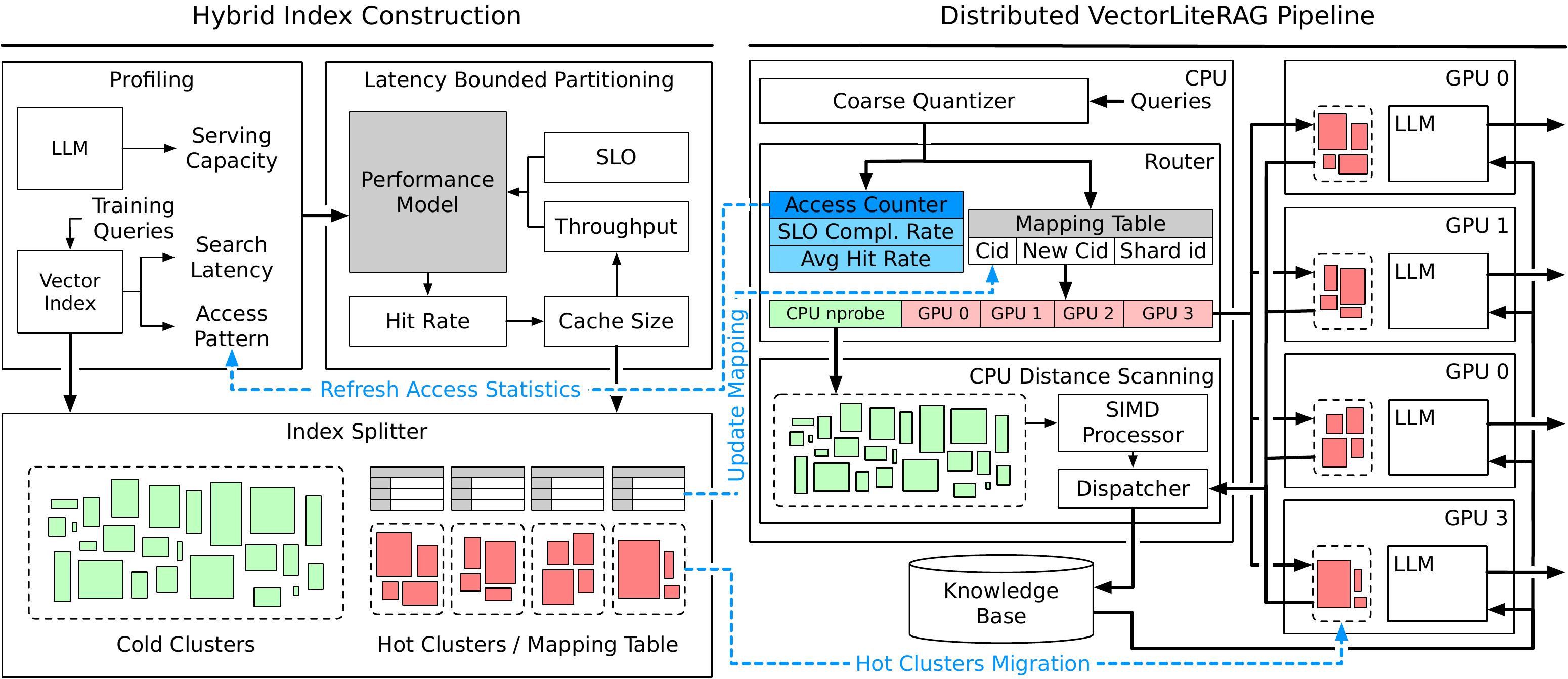} 
    \caption{System architecture of \worktitle. The system has two stages, \textbf{Left}: offline hybrid index construction and \textbf{Right}: runtime distributed pipeline. Profiling guides latency-bounded partitioning to determine cache size and split point, producing sharded indices and mapping tables. At runtime, queries are routed via coarse quantizer and mapping tables, hot clusters run on GPUs, cold clusters on CPUs. A dynamic dispatcher forwards early-finished queries to LLM workers in a timely manner. Blue trails and boxes indicate runtime index refresh and update procedures.}
    \label{fig:vectorliterag}
\vspace{-10 pt}
\end{figure*}


In summary, while GPU-based retrieval can vastly outperform CPU methods, it introduces a resource contention problem when co-located with LLMs, due to limited GPU memory and compute capacity. Meanwhile, query access patterns exhibit strong skew: a small fraction of clusters account for most retrieval traffic, making selective caching and tiered search strategies effective. However, significant variance in hit rates across queries, especially long-tail queries, poses a major challenge in latency-sensitive deployments, as batching magnifies the bottleneck introduced by slow queries. These insights motivate the design of \worktitle, which adaptively partitions the index across GPU and CPU tiers, accounting for workload skew, hit rate variance, and end-to-end latency constraints to optimize throughput and responsiveness.

\section{~\worktitle} \label{sec:overview}

\worktitle~is an optimized RAG system that determines the optimal configuration for a CPU–GPU hybrid vector index. It is organized around tightly integrated components: (1) performance modeling and latency-aware hybrid index construction, and (2) a distributed runtime pipeline for inference serving. Given the latency constraint, LLM, index, and system configuration, \worktitle~computes a partitioning point for tiered search, constructs the hybrid index, and serves inference requests through a tailored pipeline.

\textbf{Hybrid Index Construction.}
The first component of \worktitle~focuses on understanding the performance characteristics of the underlying system. This stage profiles CPU-based search latency, query-to-cluster access patterns, and standalone LLM throughput to characterize contention between retrieval and generation. These measurements drive a performance model and cache-coverage estimator, enabling a latency-bounded partitioning algorithm to select hot clusters. The hot clusters are then sharded into GPU sub-indexes.

\textbf{Distributed \worktitle~Pipeline.}
The second component is the runtime pipeline that operationalizes the hybrid index. At runtime, batched queries are routed to CPU or GPU shards using mapping tables generated during index construction, allowing each shard to operate with a flexible \texttt{nprobe} budget and reducing contention with LLM. A dynamic dispatcher further improves batching efficiency by advancing early-completing queries to mitigate tail latency.

The partitioning scheme and runtime pipeline are independent of the distance metric or compression method. As long as the index exhibits clustered structure and benefits from GPU acceleration, \worktitle~can identify an effective hybrid configuration and deliver SLO-compliant RAG service.

\subsection{Hybrid Index Construction} \label{subsec:indexconst}
\subsubsection{Profiling-based Performance Modeling}

Since GPU resources are limited, accurately modeling performance is critical for determining the optimal index partitioning point. To construct these models, \worktitle~profiles latency and access statistics using calibration queries from a training set. Specifically, it collects: (1) latency breakdown of CPU-based vector search and (2) cluster access frequency distributions. Additionally, throughput of the bare LLM is measured to guide partitioning decisions under joint CPU-GPU execution.

As described in Section~\ref{subsec:IVFsearch}, IVF index search latency is dominated by two components: coarse quantization (CQ) and LUT operations. We profile both stages across varying batch sizes and construct independent models for each. However, in our design, only the LUT stage, which corresponds to the individual distance computation and scanning step, is considered for GPU offloading for two main reasons:

First, CQ is a similarity search over the quantizer (centroid) vectors, which is often implemented using memory-intensive graph-based structures such as HNSW. Offloading CQ to GPU would require additional memory for the graph and complicate memory management.  
Second, if CQ were distributed across GPU shards, the resulting search path would involve repeated device transitions: CPU $\rightarrow$ GPU (quantization) $\rightarrow$ CPU (merge and routing) $\rightarrow$ GPU (search) $\rightarrow$ CPU (final merge). This induces costly inter-device communication and synchronization overheads. Moreover, our objective is to ensure stable performance within the latency budgets rather than to minimize absolute latency. Thus, for our purpose, we retain CQ on the CPU and use GPUs for distance computations, as this offers performance benefits while simplifying the optimization space.

Empirically, as shown in Figure~\ref{fig:searchtimetrend} (left), CPU search latency exhibits a piecewise linear relationship with batch size. Initial steps appear as the system transitions from single-threaded (single query) to multi-threaded execution (batched queries). Accordingly, we model $T^{\text{CPU}}_{\text{CQ}}$ and $T^{\text{CPU}}_{\text{LUT}}$ as piecewise linear functions of batch size.

When hot clusters are cached, the overall search time reduces accordingly. LUT operations offloaded to GPUs are fully hidden under CPU's execution, and the CPU processing time decreases in proportion to the number of hits. As a result, we model the latency of the hybrid partitioned index as:

\begin{equation}
\tau_s(b) = T^{\text{CPU}}_{\text{CQ}}(b) + (1 - \eta) \cdot T^{\text{CPU}}_{\text{LUT}}(b)
\label{eq:hybrid_latency_model}
\end{equation}

where $\eta$ denotes the hit rate, in particular the minimum hit rate among all queries in the batch.

\subsubsection{Tail Query Hit Rate Estimation}

As discussed in Section~\ref{sec:queryvar}, caching hot clusters leads to varying hit rates across queries. Because, CPU side LUT workload is proportional to the miss rate ($1 - \eta$), this variance directly translates into differences in search latency. Moreover, since vector search is typically executed in batches to maximize throughput, the completion time of the entire batch is dictated by the slowest query, one with the fewest hits. Therefore, modeling the minimum hit rate within a batch is critical for accurate performance estimation.

We model the distribution of per-query hit rates using a Beta distribution $f(x)$, which is widely used in Bayesian statistics for variables constrained to the $[0, 1]$ range. For a batch of size $b$, the expected minimum hit rate $\eta_{\min}$, i.e., the first-order statistic, is computed as:

\vspace{-1 em}
\begin{equation}
\eta_{\min}(B) = \int_0^1 B \cdot x \cdot f(x) \cdot (1 - F(x))^{B-1} \, dx
\label{eq:order_stat_min}
\end{equation}

where $F(x)$ is the cumulative distribution function of $f(x)$.

The mean hit rate $\bar{\eta}$ can be obtained directly from the query–cluster access profile, which reflects the cumulative fraction of accesses covered by the cached clusters. Estimating the variance is more challenging, as it would require re-running queries through the quantizer and counting individual hits after masking hot clusters, a process that is both computationally expensive and incompatible with iterative partitioning algorithm.

Instead, we approximate the hit rate variance as a function of the mean. We observe that hit/miss variance peaks when $\bar{\eta} = 0.5$, and becomes more uniform as $\bar{\eta} \to 0$ or $\bar{\eta} \to 1$. This mirrors the variance behavior of the Beta distribution; $\text{Var}(X) \propto \bar{\eta} (1 - \bar{\eta})$. Thus, by empirically profiling the variance at $\bar{\eta} = 0.5$, denoted $\sigma_{\max}^2$, we can approximate the variance at arbitrary $\bar{\eta}$ as: 
\[\sigma^2 \approx 4 \cdot \sigma_{\max}^2 \cdot \bar{\eta} (1 - \bar{\eta})\] 

\begin{figure} 
     \centering
     \includegraphics[width=1\linewidth]{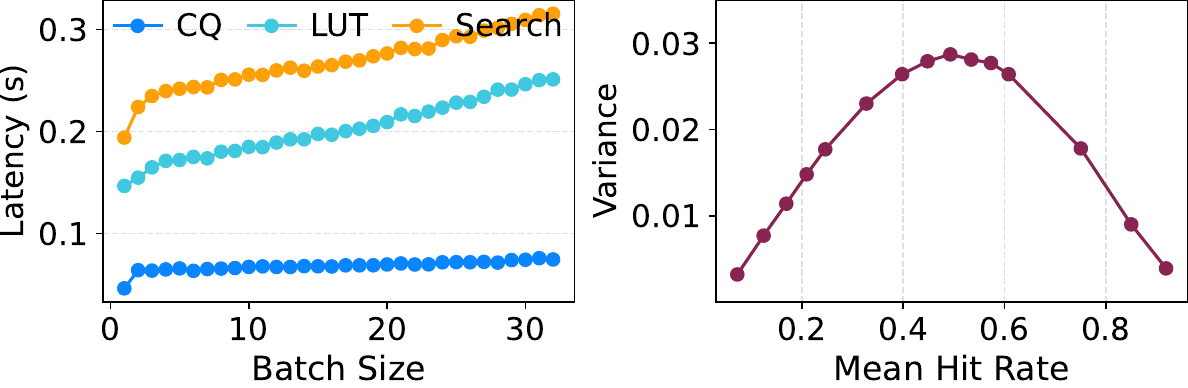} 
     \vspace{-1 em}
     \caption{\textbf{Left:} Search latency of ORCAS queries on a 64-core Intel Xeon 8426Y CPU. \textbf{Right:} Empirical variance of hit rates across queries in the Wiki-All dataset as a function of mean hit rate. The observed parabolic shape supports our variance approximation model.}
     \label{fig:searchtimetrend}
\vspace{-1 em}
 \end{figure}

Figure~\ref{fig:searchtimetrend} (right) validates the approximation. This allows instantiating a Beta distribution $f(x)$ with inferred mean and variance for any cache coverage configuration.

Finally, using Eq.~\ref{eq:order_stat_min}, we compute the minimum hit rate within a batch for a given cache coverage. Inverting this relation numerically yields the function:
\[
\rho = {HitRate2Converge}(B, \eta_{\min})
\]

which is used in the main partitioning algorithm to identify the optimal cache coverage that satisfies latency constraints.

\subsubsection{Latency-Bounded Partitioning Algorithm} \label{subsec:ppalg}

In the hybrid RAG pipeline, LLM throughput decreases as more GPU memory is allocated to the vector index, due to contention between KV cache and index storage. To balance these competing demands, we introduce an iterative algorithm that determines an index partitioning point satisfying the latency constraint.

Algorithm~\ref{alg:partition-search} outlines the proposed latency-bounded partitioning algorithm. It takes the following inputs: the latency target, the baseline KV cache memory footprint when no vector index is loaded, and the peak bare LLM throughput. The goal is to find the largest feasible cache coverage for the GPU index (partitioning point $\rho$) that satisfies SLO constraint.

We first compute the latency bound for the hybrid vector search stage. To account for queuing delay, the analysis considers a worst-case scenario in which a request arrives immediately after the previous batch begins processing. Under steady-state load with uniformly arriving requests, this tail query experiences full batch latency $W(b)$ as queuing delay. To maintain the total response time within the latency budget, the search latency must satisfy $\tau_s \le \text{SLO}_{\text{search}} - W(b)$.

To avoid circular dependency (as $W(b)$ depends on $\tau_s$), we approximate this term using a queuing factor $\epsilon$, leading to:

\begin{equation}
\tau_s = \frac{\text{SLO}_{\text{search}}}{1 + \epsilon}
\end{equation}

In our setting, we set $\epsilon = 1$, as it represents the worst case where the queuing delay equals one batch latency. This choice is empirically supported from the CPU-only baseline, where $\epsilon$ ranged between 0.9 and 1.0.

\noindent\textbf{Search iteration.} The algorithm then performs a binary search over possible values of $\rho$ using the modeled latency and hit rate behavior. For each candidate $\rho$, the reduced LLM throughput is estimated based on the corresponding decrease in KV cache capacity. Although this interpolation is coarse, it provides a conservative lower bound because the throughput–cache curve is generally convex. The \textsc{InferPartition} function is subsequently invoked to compute the expected batch size, given by $B = \mu \cdot \tau_s$, where $\mu$ is the current throughput bound. Since batch size $B$ must be an integer, two rounding strategies are considered:

- \textbf{Rounding up.} This implies longer latency and thus requires more cache coverage to meet $\tau_s$. From the hybrid latency model (Eq.~\ref{eq:hybrid_latency_model}), we solve for $\eta_1$ and convert it to coverage $\rho_1$ via the \textsc{Hitrate2Coverage} function.
  
- \textbf{Rounding down.} This yields a smaller batch size (shorter latency), but may not meet the required throughput. To ensure throughput $\mu$ is met, we solve for $\eta_2$ using the adjusted latency bound $B/\mu$ from the throughput constraint.

At the end of the iteration, the smaller of $\rho_1$ and $\rho_2$ is selected, as it requires less GPU memory. This value is used to update the binary search interval.

\vspace{0.5em}
\noindent\textbf{Convergence.}  
If the newly computed partitioning point $\rho$ increases, the resulting drop in throughput leads to a smaller batch size in the next iteration, which in turn drives $\rho$ back down. Conversely, if $\rho$ shrinks, the throughput bound increases, allowing for more cache coverage. This feedback loop ensures convergence of the algorithm within a limited number of iterations. In practice, convergence takes less than one minute as shown in Figure~\ref{fig:design_updateoverhead}.

\subsubsection{Index Splitter}

Once the partitioning point $\rho$ is determined, it is passed to the final stage of index construction, which is the index splitter. The splitter first identifies the hot clusters based on the access profile and the target cache coverage $\rho$. These hot clusters are then sorted by size and distributed to GPU shards in a round-robin fashion to balance memory usage across sub-indexes.

Alongside the construction of each sub-index, the splitter generates a set of mapping tables. These tables encode the correspondence between original cluster IDs and their assigned shard as well as the remapped local cluster IDs, enabling efficient routing during query execution.

\subsection{Distributed VectorLiteRAG Pipeline}
The right side of Figure~\ref{fig:vectorliterag} illustrates the runtime architecture of \worktitle. At initialization, memory is allocated sequentially for the index and then for the LLM to prevent memory interference between the vector search and LLM engines. The two components operate through different processes and thus use separate GPU streams for concurrency.

Similar to other IVF-based indexes, the pipeline begins with coarse quantization to identify candidate clusters. However, from this point on, \worktitle~introduces a customized retrieval pipeline tailored for hybrid CPU-GPU execution. We now describe each component in detail.


\begin{algorithm}[t]
\caption{Latency Bounded Partitioning}
\label{alg:partition-search}
\textbf{   Input:} $\text{SLO}_{\text{search}}$, $MEM_{KVcache}$, $\mu_{LLM0}$ \\
\textbf{   Output:} $\rho$
\vspace{2pt}
\begin{algorithmic}[1]
\State $\tau_s \gets \frac{\text{SLO}_{\text{search}}}{1 + \varepsilon}$
\vspace{2pt}
\State $\rho_{\text{low}} \gets 0$, $\rho_{\text{high}} \gets 1$
\While{$\rho_{\text{high}} - \rho_{\text{low}} > \delta$}
    \vspace{2pt}
    \State $\rho_m \gets \frac{\rho_{\text{low}} + \rho_{\text{high}}}{2}$
    \vspace{2pt}
    \State $\mu_{\text{LLM}} \gets \frac{MEM_{KVcache} - MEM_{Index}(\rho)}{MEM_{KVcache}}\mu_{LLM0}$
    \vspace{2pt}
    \State $\rho \gets \text{\textsc{InferPartition}}(t_s, \mu_{\text{LLM}})$
    \If{$\rho > \rho_m$}
        \State $\rho_{\text{low}} \gets \rho$
    \Else
        \State $\rho_{\text{high}} \gets \rho_m$
        \vspace{2pt}
    \EndIf
\EndWhile
\State \Return $\rho$

\State
\Function{InferPartition}{$\tau_s, \mu$}
  \State $B \gets \lceil \tau_s \cdot \mu \rceil$
  \vspace{2pt}
    \State $T^{\text{CPU}}_{\text{search}}(B), 
    T^{\text{CPU}}_{\text{LUT}}(B) \gets \text{\textsc{PerfModel}}(B)$
    \vspace{2pt}
    \State $\eta_1 \gets \frac{T^{\text{CPU}}_{\text{search}}(B)_ - \tau_s}{T^{\text{CPU}}_{\text{LUT}}(B)}$
    \State $\rho_1 \gets \text{\textsc{Hitrate2Coverage}}(\eta_1, B)$
    
  \vspace{2pt}
  \State $B \gets \lfloor \tau_s \cdot \mu \rfloor$
  \vspace{2pt}
    \State $T^{\text{CPU}}_{\text{search}}(B), T^{\text{CPU}}_{\text{LUT}}(B) \gets \text{\textsc{PerfModel}}(B)$
    \vspace{2pt}
    \State $\eta_2 \gets \frac{T^{\text{CPU}}_{\text{search}}(B) - B / \mu}{ T^{\text{CPU}}_{\text{LUT}}(B)}$
    \vspace{2pt}
    \State $\rho_2 \gets \text{\textsc{Hitrate2Coverage}}(\eta_2, B)$

\State \Return $\min(\rho_1, \rho_2)$
\EndFunction
\end{algorithmic}
\end{algorithm}


\subsubsection{Router}

To support efficient vector retrieval on a distributed multi-GPU system, \worktitle~implements a custom routing mechanism rather than relying on Faiss's built-in \texttt{IndexIVFShards}. The default implementation in Faiss is suboptimal in constrained environments for two main reasons. (1) \texttt{IndexIVFShards} partitions the index uniformly by vector or cluster ID, ignoring access frequency. While, convenient for implementation, it retains centroid metadata even for clusters that are not locally resident, causing unnecessary memory overhead, especially problematic when the number of clusters is large. (2) During search, each sub-index is instructed to probe the same number of clusters, even if many of them are not resident on that shard. Although certain probes are ultimately skipped at runtime, the batched execution of cluster scanning kernels still launches GPU thread blocks for them. These launches consume scheduling bandwidth and shared memory resources, regardless of whether the actual computation is needed. Since shared memory usage increases with \texttt{nprobe}, this results in inefficient kernel launches and exacerbates resource contention, especially in large-scale vector databases.

To address these issues, \worktitle~uses the mapping tables generated during index splitting to route each query to the appropriate GPU shards and prune irrelevant probes, thereby accounting for the device-level variance. This substantially reduces the effective \texttt{nprobe} per shard, lowering both memory pressure and kernel scheduling overhead. At runtime, only GPU workers holding relevant clusters receive and execute the search request, while the remaining portion of the search is handled by the CPU. This hybrid execution minimizes contention and enables more efficient use of GPU memory and compute resources.

\subsubsection{Dynamic Dispatcher}

Because hit rates vary across queries, the effective \texttt{nprobe} differs even within a batch. As batch size increases, the minimum hit rate tends to decrease, increasing the search latency for the entire batch. To mitigate this issue, \worktitle~employs a dynamic dispatcher that accelerates early query completion.

When search is initiated, a separate dispatcher thread is launched. Each GPU worker sets a completion flag once its assigned clusters are scanned. After all GPU flags are set, the dispatcher begins polling for queries that have completed their full search. To facilitate timely query promotion, a callback mechanism connects the CPU search loop and the dispatcher, as CPU processes clusters one-by-one, grouped by related queries. At the end of each iteration, the current scan count is compared with the expected \texttt{nprobe} for each query. When all assigned clusters for a query are scanned, the callback is invoked, and the query and its results are inserted into a thread-safe queue.

The dispatcher polls this queue at short intervals. Once a completed query is available, it merges the CPU and GPU results, re-ranks them to obtain the final top-$k$ vectors, and forwards the result to the downstream document retriever. This proactive execution reduces head-of-line blocking within batches and improves end-to-end latency, particularly for high-hit-rate queries. It also enhances batching continuity by enabling smoother transitions between retrieval and generation stages, which already employs continuous batching schemes.

\subsubsection{Adaptive Runtime Index Update}
Our model is built upon the distributional characteristics of queries aggregated across batches. While correlations among queries may temporarily shift access patterns, they primarily reduce the number of statistically independent samples rather than altering the overall distributional trend. Nevertheless, temporal bias can arise in practice, and to mitigate potential performance degradation caused by such drift, \worktitle~employs an adaptive re-profiling and update process.

\worktitle~can swiftly react to shifts in query distribution without interrupting service. During runtime, the router monitors (1) average hit rates and (2) per-cluster access frequencies. For every few minutes or after a few thousand requests, it periodically resets the counters to detect distributional drift. When the average SLO attainment falls below a threshold and observed hit rates diverge from their expected values, an update cycle is triggered: re-profiling query access patterns, rerunning the latency-bounded partitioning algorithm, generating shards, and loading the updated indices onto GPUs.

\begin{figure}
\centering
\includegraphics[width=\linewidth]{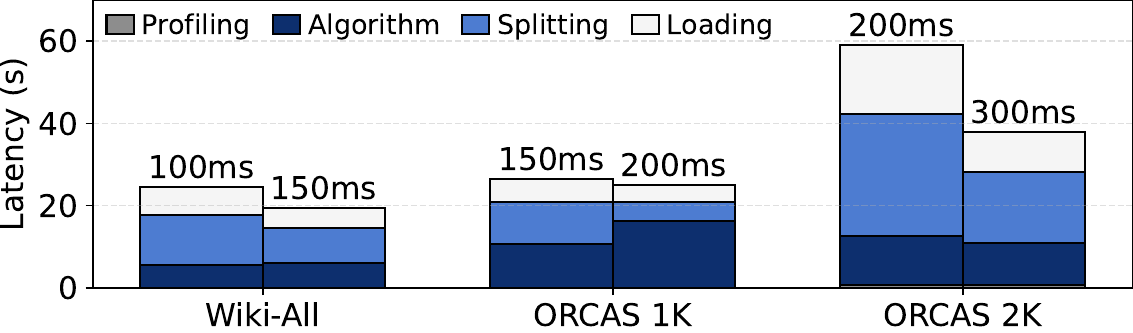}
\vspace{-2 em}
\caption{Time consumed for re-building the GPU index shards using updated query access data. Numbers above the bars denote the search time SLO constraints applied for the system.}
\label{fig:design_updateoverhead}
\vspace{-1em}
\end{figure}

All stages, from profiling to loading, complete in under a minute, allowing updates to run in the background. At the per-shard level, index generation and loading take less than ten seconds. The detailed timing breakdown for each stage is shown in Figure \ref{fig:design_updateoverhead}. While a GPU shard is being refreshed, the router temporarily redirects queries for those clusters to CPU paths, preserving the service continuity. Once the updated shard is loaded, routing automatically returns to the GPU.

Per-cluster updates are avoided because clusters are stored contiguously to enable high-bandwidth access. Since clusters vary in size, updating clusters individually would lead to memory fragmentation and inefficient data placement. Instead, \worktitle~performs full-shard updates, as migration of each shard takes only a few seconds, providing robustness and simplicity.

According to our observations, profiling with only 0.5\% of the queries from a separate training set successfully captured the distribution of 10M ORCAS queries. We therefore assume that a single index update can sustain stable service for roughly one hour under steady traffic, given the system throughput measured in our experiments.

\section{Methodology}
\subsection{Experiment Setup}
To evaluate \worktitle, we conduct experiments across various datasets, models, and hardware configurations. This section describes the datasets, models, evaluation metrics, and system setup.

\noindent\textbf{Datasets and Models.}
We use two datasets: Wiki-All and ORCAS. We construct the IVF index following the configuration guidelines provided by the Faiss library. The Wiki-All~\cite{bib:wikiall} vector database contains 88M 768-dimensional vectors derived from Wikitext~\cite{bib:wikitext} and Cohere Wikipedia embeddings, yielding a compressed IVF index with a footprint of 18GB. We also construct two additional indexes from chunked Wikipedia documents~\cite{bib:wikidump} using the Stella~\cite{bib:stella} embedding model of dimensions 1024 and 2048, and queries from the Microsoft ORCAS dataset~\cite{bib:orcas}. ORCAS consists of real Bing queries and preserves duplicates to reflect realistic query distributions. The ORCAS 1K and ORCAS 2K indexes occupy 40GB and 80GB of memory, respectively.

Our retrieval pipeline builds on Faiss v1.9.0~\cite{bib:faiss, bib:faiss_gpu}, with internal extensions for flexible \texttt{nprobe} settings and dispatcher callbacks. The overall system, including the profiler and latency-aware scheduler, is implemented in Python.

For generation, we evaluate three models—Llama3-8B, Qwen3-32B, and Llama3-70B~\cite{bib:llama3, bib:qwen3}—served using vLLM v0.9.1~\cite{bib:vllm}. The retriever and LLM run as separate subprocesses, with the main process coordinating request generation and document fetching to integrate the full RAG pipeline. 

To evaluate system performance, we sample queries from a dedicated test set that is disjoint from the profiling set. The request arrival process follows a Poisson distribution, a commonly adopted modeling choice in prior work~\cite{bib:vllm, bib:distserve, bib:splitwise}. For each query, the top-25 documents are retrieved, and a 1024-token input is constructed and passed to the LLM, which then generates a 256-token output, following the setup in~\cite{bib:hermes}. The initial \texttt{nprobe} is set to 2048, which is sufficient to achieve an average retrieval quality of 0.91 Normalized Discounted Cumulative Gain (NDCG)~\cite{bib:ndcg} at 50.

\noindent\textbf{SLO Settings.}
The SLOs for retrieval and generation stages were defined separately and then combined. For retrieval, since no standard criteria exist, we set the SLOs heuristically, relaxing them for larger databases (see Table~\ref{tab:SLO}). For generation, the SLO was defined as the latency measured at the model’s throughput limit. These capacity values were also used in building our performance model.

\begin{table}[ht]
    \centering
    \vspace{-1 em}
    \caption{SLO target values used in the main evaluation}
    \begin{tabular}{lc|lc}
        \hline
        \textbf{Vector Index } & \textbf{$SLO_{search}$} & \textbf{LLM} & \textbf{$SLO_{LLM}$} \\ 
        \hline
        Wiki-All                 & 150ms               & Llama3-8B             & 217ms               \\
        ORCAS 1K       & 200ms                & Qwen3-32B            & 191ms             \\
        ORCAS 2K       & 300ms                & Llama3-70B             & 311ms            \\
        \hline
    \end{tabular}
    \label{tab:SLO}
\end{table}


\noindent\textbf{System Configuration.}
We conduct our experiments on two types of nodes, each equipped with eight NVIDIA GPUs. The L40S node includes L40S GPUs with 48GB GDDR memory and dual Xeon 6426Y CPUs. The H100 node uses H100 GPUs with 80GB HBM and Xeon Platinum 8462Y CPUs. We use the L40S node for smaller models (Llama3-8B), while larger models requiring model parallelism (Qwen3-32B, Llama3-70B) are run on the H100 node for maximum throughput.

\noindent\textbf{Baseline Configurations.}
We compare \worktitle\ against several key baselines. Since \worktitle\ builds on FAISS, we use vanilla FAISS-CPU IVF FastScan (CPU-Only), FAISS-GPU IVF on a dedicated GPU (DED-GPU), and a sharded FAISS-GPU IVF index distributed across all GPUs (ALL-GPU).
To further demonstrate the strength of our approach, we also compare against HedraRAG~\cite{bib:hedrarag} in section \ref{subsec:hedraRAG}, which also uses a skew-aware caching strategy. 
\section{Evaluations}
\subsection{Performance Model and Hit Rate Estimator}

\begin{figure}
\centering
\includegraphics[width=\linewidth]{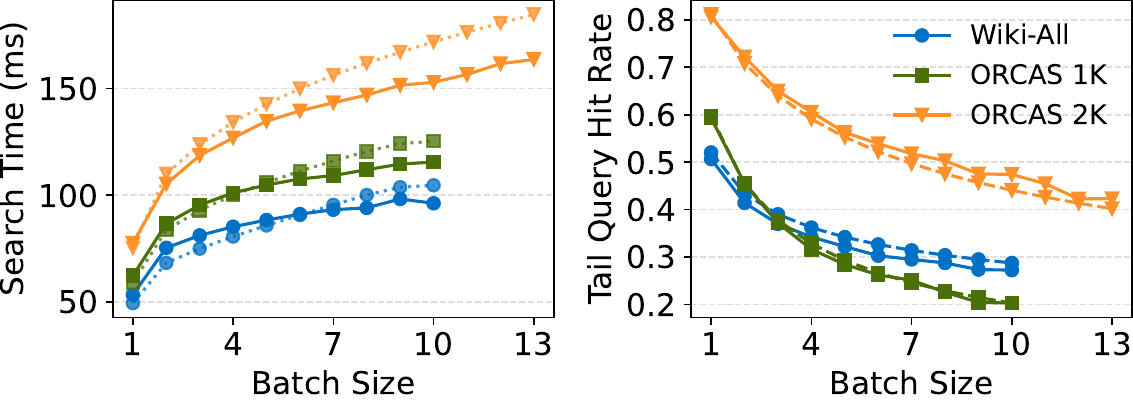}
\vspace{-2 em}
\caption{Comparison of measured (solid line) vs estimated (dotted line) values from \textsc{VectorLiteRAG}’s performance model. \textbf{Left:} Search latency across batch sizes.
\textbf{Right:} Tail hit rates within a batch.}
\label{fig:perfmodelresult}
\vspace{-1 em}
\end{figure}

Figure~\ref{fig:perfmodelresult} evaluates the accuracy of \vectorlite's performance model. The right panel compares the predicted and actual minimum hit rates within each batch. As expected from order statistics, the minimum hit rate declines rapidly as batch size increases, and the rate of decline gradually flattens in the large-batch regime. Close alignment of two curves confirms that our Beta-distribution-based approximation reliably captures caching effectiveness.

The left panel compares the predicted latency of the hybrid index search with the measured latency. While the predictions generally follow the same trend, a noticeable offset exists between the two. This deviation mainly results from the dispatcher’s early-query handling, as discussed in Section~\ref{subsec:dispatcher}. 

 Precisely capturing the dispatcher’s impact would require evaluating full order statistics to model per-request completion times, which greatly increases complexity while providing only marginal benefit. Despite these approximations, the resulting configurations perform robustly in practice, as shown in the following sections.

\begin{figure*}
\centering
\includegraphics[width=\linewidth]{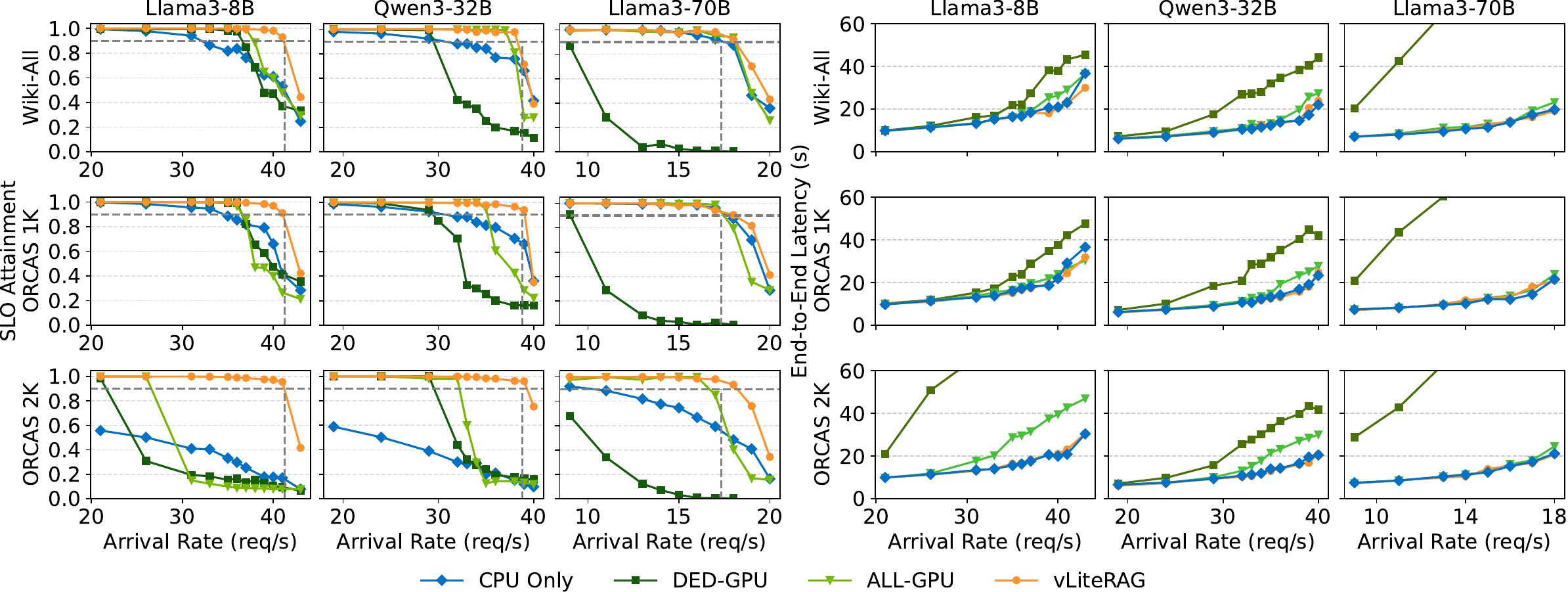}
\vspace{-2 em}
\caption{\textbf{Left}: TTFT SLO attainment and \textbf{Right}: end-to-end latency of RAG pipeline under increasing arrival rates across different LLMs (columns) and datasets (rows). Our work (vLiteRAG) achieves higher SLO attainment across all regimes compared to baselines.}
\label{fig:main_eval}
\vspace{-1 em}
\end{figure*}

\subsection{SLO Attainment}
Figure~\ref{fig:main_eval} presents SLO attainment curves across all nine combinations of vector databases and LLMs. In each subplot, the horizontal dashed line marks the 90th percentile latency target, and the vertical dashed line indicates the standalone LLM throughput. All experiments use on-demand dynamic batching, where retrieval requests are served immediately after the previous search completes, allowing throughput to scale with arrival rate through adaptive batch sizing.

Across all configurations, \worktitle~sustains the extended SLO budget ($SLO_{\text{LLM}} + SLO_{\text{Search}}$, defined in Table~\ref{tab:SLO}) over the widest input rate ranges among evaluated baselines.  CPU-based fast scan can support relatively high request per second (RPS) rates, its limited per-request performance leads to consistent SLO violations even under light traffic. As arrival rate increases, batch sizes grow (up to 9–10 under $>$40 RPS), incurring high latency and poor tail response.

Dedicated GPU retrieval performs poorly with large models due to rigid model parallelism constraints. For instance, Llama3-70B requires a tensor parallelism degree of 4 for efficient execution. While it fits within 2 H100 GPUs, the achievable LLM throughput drops from ~8 RPS to less than 2 RPS. In such settings, dedicating GPU(s) to retrieval results in resource oversubscription, harming overall system throughput.

For small vector databases and under light loads, ALL-GPU configurations can satisfy SLOs over wide traffic ranges. However, as the arrival rate approaches its reduced throughput, latency increases sharply. Although \worktitle~is subject to this limitation as well, its optimized partitioning algorithm extends the SLO-attainable region nearly up to the standalone LLM throughput limit.

\begin{figure}
\centering
\includegraphics[width=\linewidth]{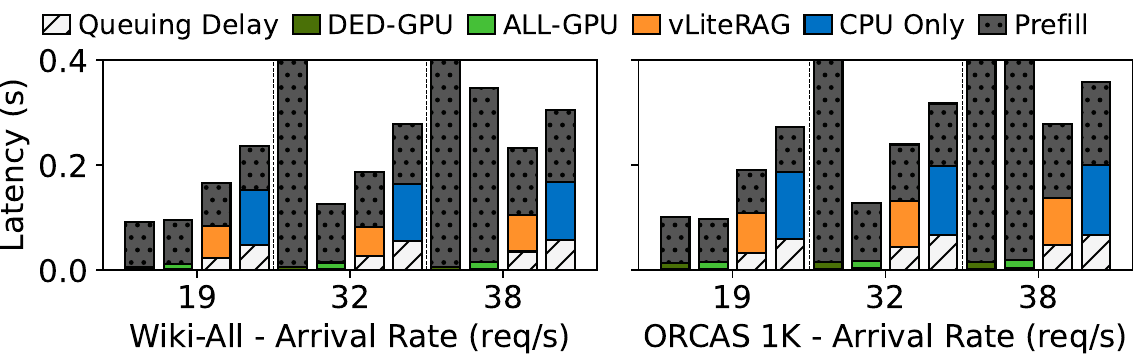}
\vspace{-2 em}
\caption{TTFT breakdown for Wiki-All and ORCAS 1K indexes with Qwen3-32B. Each group shows results from four configurations. Bars are stacked to show the contribution of queuing delay, vector search latency (colored segments), and LLM prefill latency (grey)}
\label{fig:breakdown}
\vspace{-1 em}
\end{figure}

To better illustrate the dynamics of RAG systems, we present a detailed TTFT breakdown in Figure~\ref{fig:breakdown} for the Qwen3-32B model with Wiki-All and ORCAS 1K indices under varying input rates. As search latency increases, especially with CPU-based retrieval, queuing delays compound, further inflating TTFT. While both dedicated and ALL-GPU shared baselines perform well under low traffic, they exhibit latency spikes at higher rates due to resource contention. In contrast, \worktitle~sustains stable latency by balancing throughput and latency, enabling finer control over resource allocation across the RAG stages.

\subsection{End-to-End Latency}
Since GPU resources are shared between retrieval and generation, interference with the decoding phase is inevitable. To assess the impact of such interference, we present the end-to-end latency results from the nine configurations discussed earlier, shown in Figure~\ref{fig:main_eval}. 

Retrieval contention is most severe for smaller models that can sustain higher loads, whereas large models saturate compute resources before retrieval pressure dominates. In the low-traffic regime, contention is minimal, except in DED-GPU, which reduces the number of GPUs available to the LLM. However, under high traffic and with large vector databases, contention becomes significant. This is evident in the more than 2$\times$ increase in end-to-end latency observed in ALL-GPU baselines for ORCAS 2K with Llama3-8B and Qwen3-32B. Although Llama3-70B involves more intensive computation, its low throughput ceiling causes TTFT to diverge before retrieval-induced interference becomes the dominant factor.

In contrast, \worktitle~matches CPU-based retrieval in end-to-end latency. This demonstrates that its partitioning strategy and distributed execution pipeline effectively minimizes interference by carefully limiting GPU memory and usage of GPU threads for retrieval, thereby preserving LLM generation performance, while maintaining latency lower than SLO requirements.

\begin{figure}
\centering
\includegraphics[width=\linewidth]{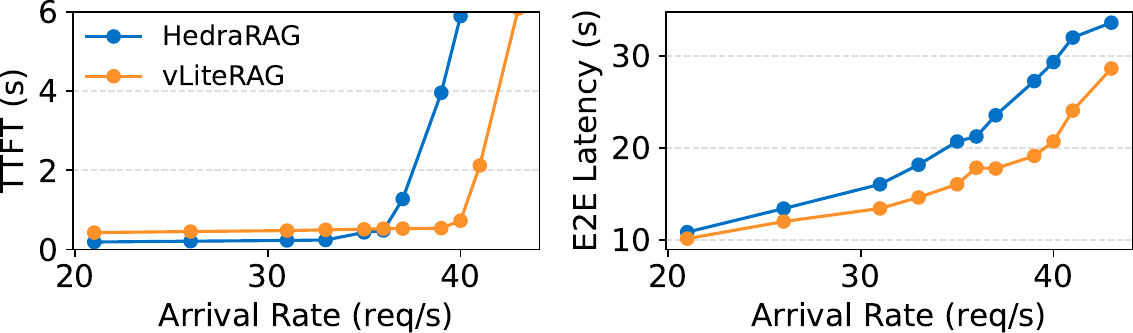}
\vspace{-2 em}
\caption{Comparison with HedraRAG. HedraRAG exhibits lower TTFT at low request rates, but latency increases sharply once the system exceeds its throughput limit. \worktitle~ is configured with $SLO_{search} = 400$ms.}
\label{fig:hedraRAG}
\vspace{-1 em}
\end{figure}

\vspace{-0.1 em}
\subsection{Comparison with HedraRAG} ~\label{subsec:hedraRAG}
\vspace{-1 em}

We compare \worktitle~with HedraRAG\cite{bib:hedrarag}, which also exploits skewed cluster access patterns in RAG pipelines. While both systems adopt tiered caching strategies for vector indices, their partitioning principles and target objectives differ fundamentally.

HedraRAG selects GPU-resident clusters by identifying the maximum KV cache size that can sustain the throughput of the slower stage, either the LLM or the retriever. Although this approach is simple and throughput-aware, it does not account for latency constraints that are critical for real-time serving. In configurations where the LLM stage exhibits lower peak throughput than retrieval, as in~\ref{fig:main_eval}, HedraRAG allocates the entire GPU memory to LLMs and performs vector search on the CPU. As noted in their paper, HedraRAG is most effective when retrieval becomes extremely heavy.

To enable a fair comparison, we replicate the HedraRAG setting by building an IVF index with $\sqrt{N_{\text{vector}}}$ clusters and measuring retrieval throughput using batch sizes below 64. At $\text{\texttt{nprobe}}=256$, CPU-only retrieval achieves 35 RPS at 0.94 NDCG@50; we increase \texttt{nprobe} to 6144 in our system to match this accuracy. Since HedraRAG does not support distributed retrieval, we apply their GPU caching scheme using \texttt{IndexIVFShard} without our optimized pipeline.

Figure~\ref{fig:hedraRAG} summarizes the results. HedraRAG places 73\% of index clusters in GPU memory, whereas \worktitle~identifies a partitioning point of 31.5\% under a 400,ms SLO. While HedraRAG achieves lower retrieval latency under low traffic, its operable range narrows as input rates increase. In contrast, \worktitle~maintains latency near the target constraint across a wider traffic range and achieves lower overall end-to-end latency through its distributed pipeline.

The key distinction lies in how partitioning decisions are made. \worktitle~allows operators to specify a target SLO and computes the largest GPU-resident index region that satisfies this constraint, whereas HedraRAG balances throughput between stages without explicit latency objectives, which can lead to suboptimal GPU allocation.

\subsection{Ablation Studies}
\subsubsection{Dynamic Dispatcher} 
\label{subsec:dispatcher}

\begin{figure}
\centering
\includegraphics[width=\linewidth]{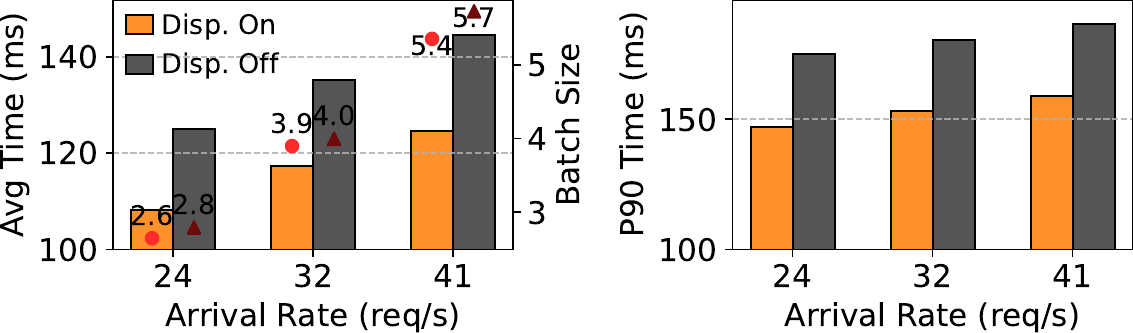}
\vspace{-1 em}
\caption{\textbf{Left:} Average search latency and batch sizes. \textbf{Right:} P90 tail latency on ORCAS 2K index with dispatcher enabled and disabled.}
\label{fig:dispatcher}
\vspace{-1 em}
\end{figure}

Figure~\ref{fig:dispatcher} illustrates the effectiveness of the dynamic dispatcher in the distributed \worktitle~pipeline. By polling the scanning loop and dispatching queries immediately upon completion, the dispatcher reduces search latency by up to 16\%, improving both average and tail latency. This gain is achieved by overlapping the merging and re-ranking of early-completed queries with the ongoing scanning of slower queries, avoiding bulk merging at the end.

Figure~\ref{fig:dispatcher} also reports average batch sizes under varying arrival rates. With adaptive batching, requests are grouped dynamically based on current pipeline load. Since vector search has higher throughput capacity than the LLM, it absorbs higher arrival rates by increasing batch size while maintaining stable service time. In contrast, fixed or capped batch sizes lead to request backlogs and performance degradation.

\subsubsection{Impact of LLM Input and Output Lengths}

\begin{figure}
\centering
\includegraphics[width=\linewidth]{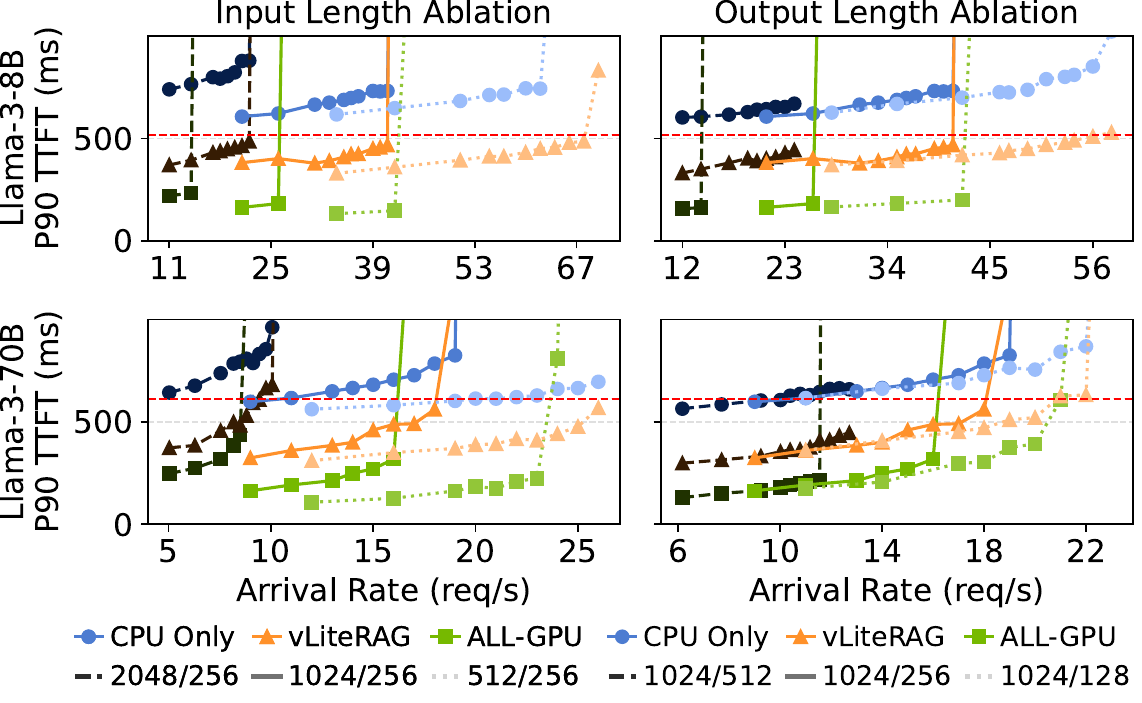}
\vspace{-1 em}
\caption{\textbf{Left}: P90 TTFT across different input and \textbf{Right}: output lengths. Darker curves represent longer input/output sequences, while brighter curves correspond to shorter ones. Experiments were conducted using the ORCAS-2K index.}
\label{fig:abl_inoutlength}
\vspace{-1 em}
\end{figure}

Figure~\ref{fig:abl_inoutlength} illustrates latency sensitivity to varying input and output lengths for Llama3-8B and Llama3-70B. The red dashed line denotes the combined SLO target of vector search and LLM stages, corresponding to the 1024/256 setting in Table~\ref{tab:SLO}. For consistency, $\text{SLO}_{LLM}$ is fixed across configurations.

Longer inputs increase prefill cost, raising TTFT and shifting SLO violations to lower arrival rates as compute resources saturate. Similarly, longer outputs reduce the SLO-compliant range due to extended generation time and higher KV cache usage. Across both dimensions, \worktitle~maintains serviceability over a wider range than the baselines, highlighting the robustness of its partitioning scheme.

\subsubsection{Sensitivity study on \texorpdfstring{$\text{SLO}_{search}$}{SLOsearch}}

\begin{figure}
\centering
\includegraphics[width=\linewidth]{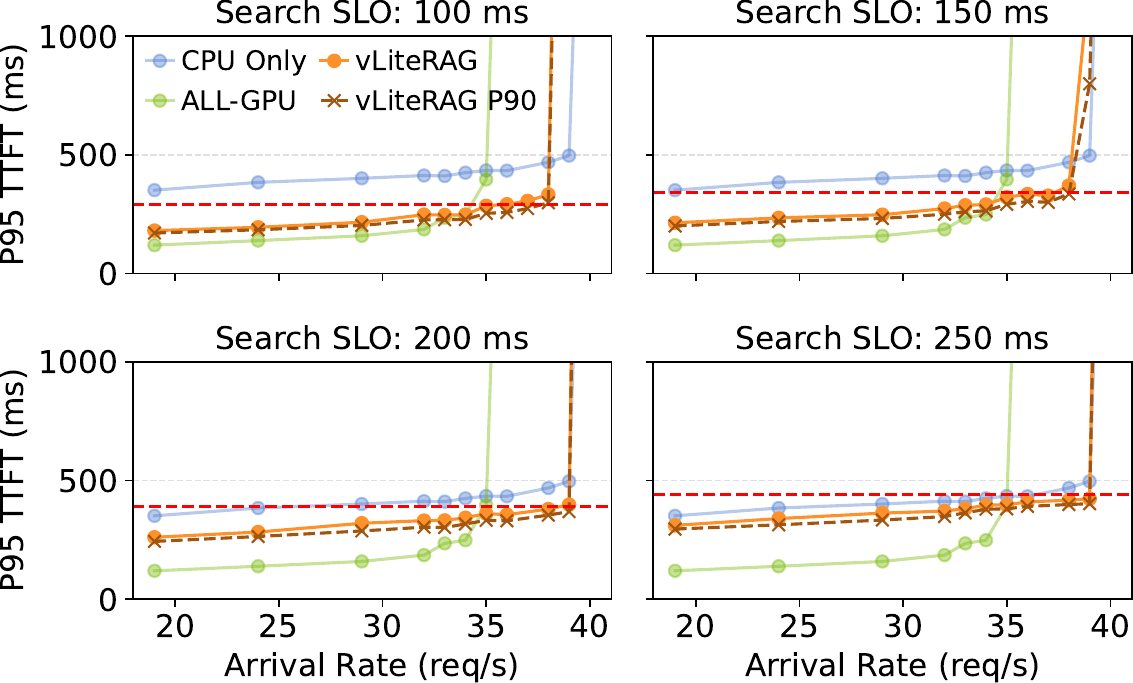}
\vspace{-1 em}
\caption{P95 tail latency (and P90 for \worktitle) under different search-stage SLO constraints. Results are obtained using the Qwen3-32B model and the ORCAS 1K index.}
\label{fig:abl_slo}
\vspace{-1 em}
\end{figure}

To evaluate the robustness of our system under varying service constraints, we test \worktitle~across multiple $\text{SLO}_{search}$ targets. All plots in Figure\ref{fig:abl_slo} use P95 TTFT as the primary metric, with P90 results additionally shown as dashed lines for \worktitle. Changing the quantile slightly expands or shrinks the SLO-compliant range; in our evaluation, the difference between P90 and P95 was at most 1~RPS.

\begin{table}[H]
\centering
\vspace{-1 em}
\caption{SLO targets and corresponding index shard sizes.}
\begin{tabular}{c|c|c|c}
\hline
\textbf{SLO (ms)} & \textbf{Index (GB)} & \textbf{Param (GB)} & \textbf{KV Cache (GB)} \\
\hline
100 & 3.80 & \multirow{4}{*}{30.59} & 33.24 \\
150 & 2.95 &                       & 34.09 \\
200 & 2.47 &                       & 34.57 \\
250 & 2.21 &                       & 34.83 \\
\hline
\end{tabular}
\vspace{-1 em}
\label{tab:slo_index_cache}
\end{table}

Table~\ref{tab:slo_index_cache} summarizes the target SLOs and their associated memory allocations. Under relaxed SLO constraints, the latency-bounded partitioning algorithm assigns a smaller fraction of the index to GPU shards, yielding latency behavior closer to the CPU-only baseline. As the SLO becomes stricter, the latency curve moves toward the all-GPU configuration. While tighter SLOs reduce available KV-cache space and modestly shrink the operable region, \worktitle~still delivers a wider SLO-compliant throughput range than the baselines, highlighting the adaptability of its partitioning strategy and the effectiveness of its execution pipeline.

\subsubsection{Robustness to Hardware Capacity}

Finally, we evaluate how \worktitle~adapts to different hardware capacities of the system. Following the provisioning policy commonly adopted by cloud providers, which allocates additional CPU cores as more GPUs are added, we test three configurations: 4 GPUs + 32 cores, 6 GPUs + 48 cores, and 8 GPUs + 64 cores. For each configuration, we re-profile the CPU-only search latency and apply the same latency-bounded partitioning algorithm. Aside from the number of compute devices, all experiments use identical model and index setups.

\begin{figure}
\centering
\includegraphics[width=\linewidth]{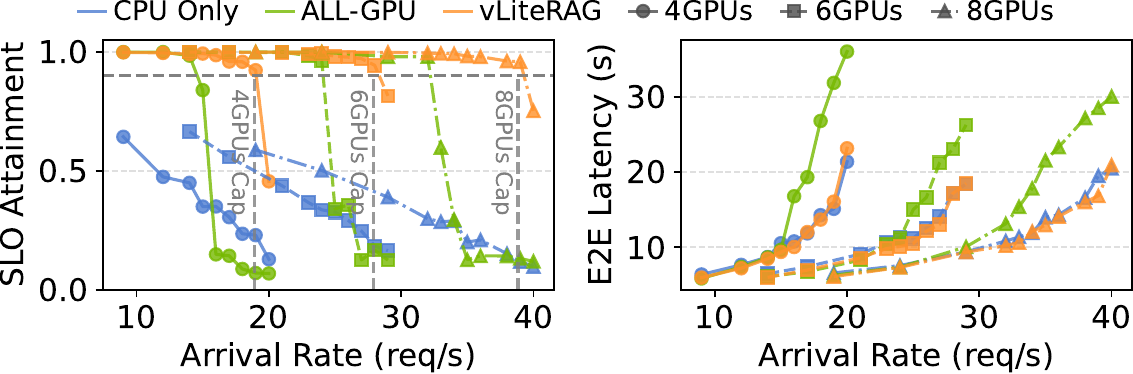}
\vspace{-1 em}
\caption{\textbf{Left:} SLO attainment (the vertical dashed line denotes bare LLM capacity) and \textbf{Right:} end-to-end latency measured on 4-, 6-, and 8-GPU systems. Evaluated using the Qwen3-32B model and the ORCAS 2K index.}
\label{fig:abl_gpunum}
\vspace{-1 em}
\end{figure}

The results in Figure~\ref{fig:abl_gpunum} show that \worktitle~sustains the target SLO across all configurations while extending the SLO-compliant throughput roughly in proportion to the number of GPUs. While the reduced memory capacity in the GPU baseline causes decoding latency to grow rapidly with scale, \worktitle~effectively contains this growth, keeping decoding latency comparable to CPU-only search cases. This demonstrates that \worktitle~can be readily deployed across clusters of different sizes with minimal setup effort while maintaining consistent latency behavior.


\section{Related Works}

RAG applications with iterative retrieval or multi-stage generation often exhibit semantic similarity across successive queries. Motivated by this observation, several optimization techniques have been proposed, including prefetching~\cite{bib:telerag}, speculative retrieval~\cite{bib:specrag}, and pipelined execution~\cite{bib:piperag}. In contrast, our work builds upon application-agnostic, generic retrieval–generation pipelines without relying on semantic priors or intermediate signals. RagCache~\cite{bib:ragcache} improves throughput by managing KV cache reuse between tenants, focusing on scheduling and reuse optimizations on the LLM side. Hermes~\cite{bib:hermes}, on the other hand, scales via disaggregation by adding CPU nodes to offload vector search.

Efforts such as~\cite{bib:anna, bib:chameleon, bib:cxl-anns, bib:accelrag, bib:instoragerag} propose specialized hardware or memory-centric architectures to accelerate RAG pipelines. While these approaches offer significant performance gains, they often rely on custom infrastructure, which may limit deployability in general-purpose environments. Among prior works, HedraRAG~\cite{bib:hedrarag} also co-locates retrieval and generation on GPUs. Our work builds on this direction with an analytical model for latency and hit rate, enabling principled GPU memory partitioning under explicit SLOs. To our knowledge, \worktitle~is the first solution to provide fine-grained resource control for co-located RAG pipelines.

Future work may extend our approach to prefill–decode disaggregation frameworks~\cite{bib:distserve,bib:splitwise}, where bandwidth-bound retrieval may run alongside compute-intensive prefill. This would require jointly modeling vector search and the throughput of both stages, but our framework offers a natural basis for such integration.

\vspace{-3 pt}

\section{conclusion}

This paper presents \worktitle, a latency-aware orchestration framework for Retrieval-Augmented Generation (RAG) systems that explicitly manages the tight coupling between vector retrieval and LLM inference. We show that under skewed access patterns, variability in retrieval latency interacts with inference batching, causing tail effects amplification that cannot be mitigated by optimizing either stage in isolation.

\worktitle~is driven by the insight that meeting strict RAG SLOs requires balancing batching behavior rather than maximizing instantaneous GPU utilization. By coordinating retrieval progress with inference scheduling, \worktitle~suppresses tail cascades and sustains predictable end-to-end latency under bursty workloads. This enables SLO compliance across a substantially wider operating regime, supporting up to 1.5× higher request rates than baseline RAG systems.
Across extensive evaluation, we demonstrate that these benefits generalize across latency targets, hardware configurations, and LLM input/output lengths. \worktitle~further exposes explicit control knobs that allow RAG operators to trade throughput for tail latency under constrained GPU memory budgets, making it practical for real-world deployment.

\section{Acknowledgment}

This research was supported in part through cyber-infrastructure research resources and services provided by the Partnership for an Advanced Computing Environment (PACE) at the Georgia Institute of Technology, Atlanta, Georgia, USA.
This work was partially supported by gifts from Google and AMD.
The views and conclusions contained herein are those of the authors and should not be interpreted as representing the official policies or endorsements, either expressed or implied, of Georgia Tech.
\appendix

\subsection{Abstract}
The artifact includes the complete source code of the core \worktitle~system, together with our modified FAISS library used for hybrid CPU–GPU vector search. To ensure reproducibility of both preprocessing and evaluation, we also provide a collection of shell scripts and Python utilities that automate the full experimental workflow, including dataset preparation, index construction, performance profiling, and end-to-end RAG pipeline evaluation. These scripts are designed to reproduce all major results reported in the paper with minimal manual intervention. 

All code and supporting materials are publicly available on GitHub 
\href{https://github.com/sitar-lab/VectorLiteRAG-AE.git}{https://github.com/sitar-lab/VectorLiteRAG-AE} and Zenodo \href{https://zenodo.org/records/18195323}{https://zenodo.org/records/18195323}

\subsection{Artifact Check-list}
\begin{itemize}
    \item \textbf{Program:} Modified FAISS library, vLLM
    \item \textbf{Compilation:} \texttt{gcc-11.3}, \texttt{nvcc-12.1}, \texttt{cmake}.
    \item \textbf{Models:} Llama-3 8B, Llama-3 70B, and Qwen-3 32B.
    \item \textbf{Datasets:} MS ORCAS and NVIDIA Wiki-All.
    \item \textbf{Run-time Environment:} RHEL 9 with Anaconda3.
    \item \textbf{Hardware:} Single node equipped with 8 NVIDIA L40S GPUs and 8 NVIDIA H100 GPUs.
    \item \textbf{Metrics:} SLO attainment, end-to-end latency, vector search hit rate estimation.
    \item \textbf{Output:} CSV logs and visualization plots.
    \item \textbf{Disk Space Required:} \(\sim\)256 GB for evaluation; \(\sim\)1.5 TB for preprocessing and index construction.
    \item \textbf{Workflow Preparation Time:} 40--50 hours.
    \item \textbf{Experiment Completion Time:} 10 hours.
    \item \textbf{Publicly Available?:} Yes.
    \item \textbf{Code Licenses?:} CC BY 4.0
    \item \textbf{Archived(DOI)?:} \href{https://zenodo.org/records/18195323}{https://zenodo.org/records/18195323}
\end{itemize}

\subsection{Description}
\subsubsection{How to access} All source cod and scripts are accessible via github repository.

\subsubsection{Hardware dependencies} All experiments were conducted on a single L40S node or a single H100 node, each equipped with 8 GPUs. Because larger language models rely on tensor model parallelism, the H100 system is expected to provide NVLink connectivity to ensure reproducible performance. The L40S node was configured with a 32-core Intel CPU, and the H100 node with a 64-core Intel CPU. CPU core count is an important factor, as a substantial portion of the workload executes on the host processor.

\subsubsection{Software dependencies} The evaluation environment was run on RHEL 9 (or a compatible Linux distribution) using Anaconda3. Successful compilation of the FAISS library depends on specific toolchain versions, including Python 3.10, GCC 11.3, and NVCC 12.1. Intel MKL is also required to support vectorized CPU operations.

\subsubsection{Datasets} The Wikiall benchmark is directly downloadable.
The ORCAS 1K and ORCAS 2K benchmarks require both the MS ORCAS dataset and the English Wikipedia dump, which are publicly accessible but require long preprocessing.
\subsection{Installation and Testing}
\subsubsection{Installation}
{\footnotesize
\begin{verbatim}
  # Create conda environment
  cd VectorLiteRAG
  conda create -n vlite -f ./scripts/env.yml
  conda activate vlite

  # Build faiss library
  git submodule update --init --recursive
  ./scripts/build.sh
\end{verbatim}}

\subsubsection{Preprocessing}
{\footnotesize
\begin{verbatim}
  # Download a samll dataset for testing
  ./database/download.sh test

  # Chunk documents and run embedding model
  ./database/encode.sh test

  # Train index and construct base IVF
  ./scripts/train.sh test
\end{verbatim}}

\subsubsection{Testing}
{\footnotesize
\begin{verbatim}
  # Run a round of short cpu search based RAG
  # Output csv files will appear under results/test
  ./scripts/test.sh
\end{verbatim}}

\subsection{Experiment Workflow}
We provide scripts for evaluation along with corresponding plotting utilities. The workflow is straightforward. After completing the preprocessing steps, execute the following commands from the project’s root directory:
\begin{itemize}
\item[(1)] Download the datasets:
{\small\verb|./database/download.sh|}
\item[(2)] Perform preprocessing and train the base indexes:  \\
    {\small\verb|./database/encode.sh <dataset>|}  \\
    {\small\verb|./scripts/train.sh <dataset>|}

\item[(3)] Run all experiments sequentially: \\  
    {\small\verb|./scripts/runall.sh|}  \\
    For running experiments individually, please refer to the repository README.

\item[(4)] After all experiments are completed, generate the figures:  
    {\small\verb|./scripts/plotall.sh|}  \\
    Individual plotting options are also documented in the README.
\end{itemize}

\subsection{Evaluation and Expected Results}
This artifact reproduces the primary experimental results presented in Figures 10–17 of the paper. During evaluation, latency logs are saved under \texttt{results/<datasets>}, and all vector indexes and their associated metadata are stored in \texttt{database/<dataset>}. After the evaluation completes, the provided plotting scripts generate the corresponding figures and place them in \texttt{figures} directory.

The reproduced results are expected to closely align with those reported in the paper, though minor variations may occur due to system-level factors. Individual evaluation runs for each data point are also supported pyand documented in the repository, enabling users to verify and assess any deviation.


\balance
\bibliographystyle{IEEEtranS}
\bibliography{refs}

\end{document}